\documentclass[letterpaper]{article} 
\usepackage{aaai2026}  
\usepackage{times}  
\usepackage{helvet}  
\usepackage{courier}  
\usepackage[hyphens]{url}  
\usepackage{graphicx} 
\usepackage{natbib}  
\usepackage{caption} 
\usepackage[linesnumbered,ruled,vlined]{algorithm2e}
\usepackage{cite}
\usepackage{amsmath,amssymb,amsfonts}
\usepackage{multirow}
\usepackage{bm}

\usepackage{tikz}
\usetikzlibrary{shapes}
\usetikzlibrary{positioning}
\usetikzlibrary{plotmarks}
\usetikzlibrary{patterns}
\usetikzlibrary{fit,calc}
\usetikzlibrary{arrows.meta}
\usepackage{microtype}
\usepackage{graphicx}
\usepackage{enumitem}
\usepackage{booktabs}
\usepackage{adjustbox}
\usepackage{xspace}
\usepackage{multirow}
\usepackage{tabularx}
\usepackage{tikz}
\usepackage{subcaption}
\usepackage{tikz}
\usetikzlibrary{positioning}
\usepackage{thmtools, thm-restate}
\usepackage{colortbl}
\usepackage{threeparttable}
\usepackage{pifont}

\usepackage{newfloat}
\usepackage{listings}
\DeclareCaptionStyle{ruled}{labelfont=normalfont,labelsep=colon,strut=off} 
\newcommand{\g}[1]{\textcolor{gray!50}{#1}} 

\nocopyright 

\title{Probing then Editing: A Push-Pull Framework for Retain-Free \\ Machine Unlearning in Industrial IoT}
\author {
    Jiao Chen\textsuperscript{\rm 1},
    Weihua Li\textsuperscript{\rm 2},
    Jianhua Tang\textsuperscript{\rm 1 3 \footnote{Corresponding author.}}
}

\affiliations{
202110190459@mail.scut.edu.cn, whlee@scut.edu.cn, jtang4@e.ntu.edu.sg
}

\usepackage{bibentry}

\begin{document}
\maketitle

\footnotetext[1]{Shien-Ming Wu School of Intelligent Engineering, South China University of Technology, Guangzhou, China.}
\footnotetext[2]{School of Mechanical and Automotive Engineering, South China University of Technology, Guangzhou, China.}
\footnotetext[3]{Key Laboratory of Cognitive Radio and Information Processing, Ministry of Education (Guilin University of Electronic Technology), Guilin, China.}

\begin{abstract}
In dynamic Industrial Internet of Things (IIoT) environments, models need the ability to selectively forget outdated or erroneous knowledge. However, existing methods typically rely on retain data to constrain model behavior, which increases computational and energy burdens and conflicts with industrial data silos and privacy compliance requirements. To address this, we propose a novel retain-free unlearning framework—Probing then Editing (PTE). PTE frames unlearning as a ``probe-edit" process: first, it probes the decision boundary neighborhood of the model on the to-be-forgotten class via gradient ascent and generates corresponding editing instructions using the model's own predictions. Subsequently, a push-pull collaborative optimization is performed: the ``push" branch actively dismantles the decision region of the target class using the editing instructions, while the ``pull" branch applies masked knowledge distillation to anchor the model's knowledge on retained classes to their original states. Benefiting from this mechanism, PTE achieves efficient and balanced knowledge editing using only the to-be-forgotten data and the original model. Experimental results demonstrate that PTE achieves an excellent balance between unlearning effectiveness and model utility across multiple general and industrial benchmarks such as CWRU and SCUT-FD.
\end{abstract}

\section{Introduction}
\label{sed:introduction}
Deep neural networks have become the cornerstone of modern industrial intelligence, driving critical applications such as predictive maintenance, quality control, and manufacturing process optimization~\cite{10697107,ren2024industrial,our_comst,li2022perspective}. However, Industrial Internet of Things (IIoT) environments are inherently dynamic~\cite{10018896,10102331}. Changes in raw material batches, updates to production protocols, or sensor degradation can render certain operational data categories outdated or irrelevant~\cite{ren2025industrial}. This necessitates models not only to learn new knowledge but also to selectively forget stale or incorrect knowledge. Such capability is crucial for sustainable and efficient model lifecycle management, avoiding the high computational and energy costs of full retraining, aligning with the principles of Green AI~\cite{10637271}.

\begin{figure}[t]
    \centering
    \includegraphics[width=0.9\linewidth]{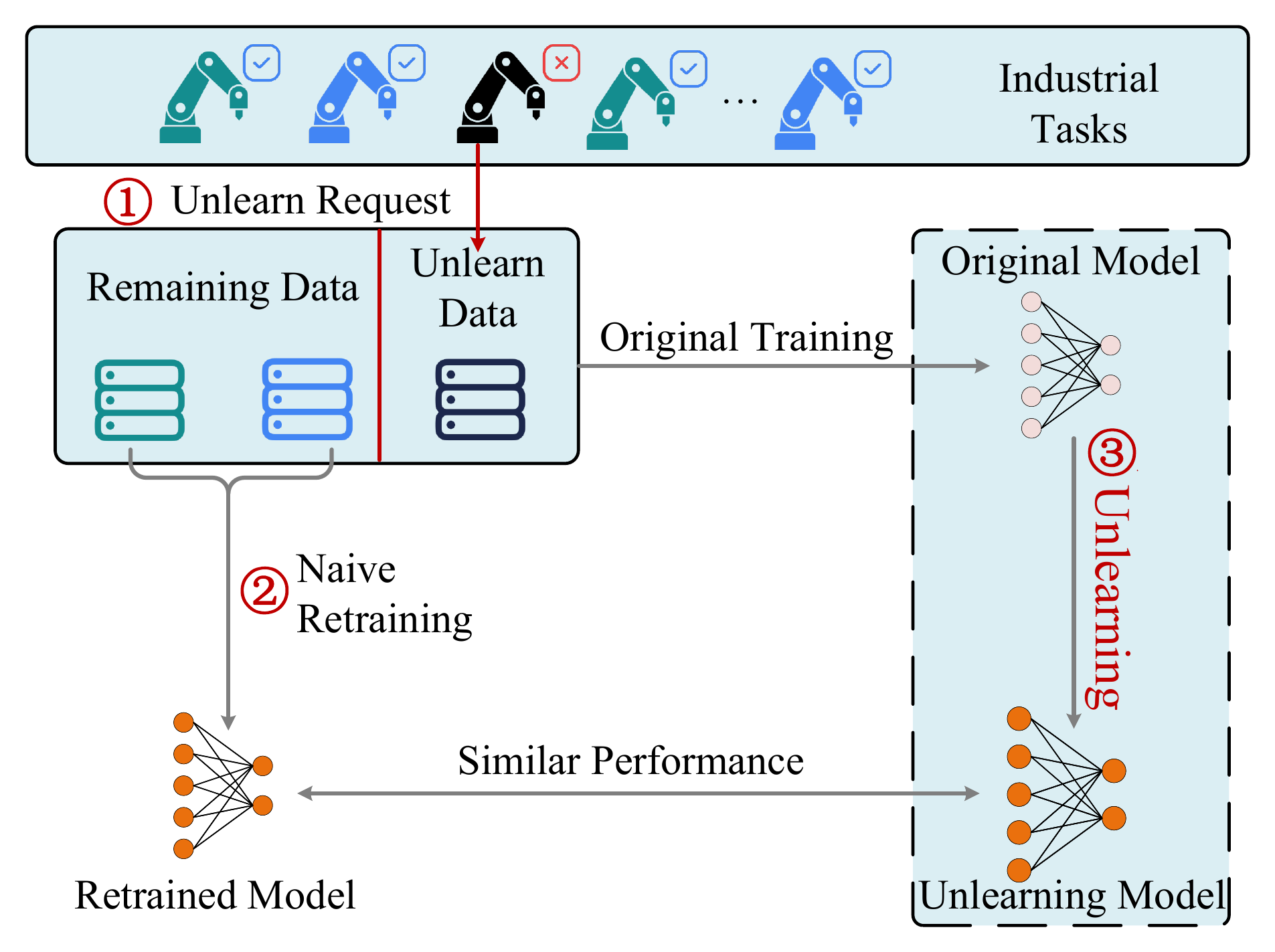}
    \caption{Machine Unlearning and Naive Retraining in Industrial IoT Scenarios}
    \label{fig:unlearning_process}
\end{figure}

Machine unlearning~\cite{li2025machine,nguyen2022survey,shaik2024exploring} offers an ideal solution by aiming to remove the influence of specific data subsets from a trained model, making it behave as if it had never learned that data. 
As illustrated in Figure~\ref{fig:unlearning_process}, a typical unlearning process begins with a model trained on the full dataset, including both retained and removable data. Upon receiving a \textbf{\ding{172} unlearn request} - which may be triggered by privacy concerns, dataset updates, or erroneous labels - the naive strategy is to \textbf{\ding{173} retrain} the model from scratch using only the retained data. Although this guarantees complete data removal, it incurs prohibitive computation and energy costs. Alternatively, \textbf{\ding{174} machine unlearning} seeks to achieve the same outcome more efficiently by adjusting the model directly to eliminate the influence of the removed data while preserving performance in the retained dataset.

Early efforts focused on exact unlearning, where the unlearned model is mathematically equivalent to one retrained on the remaining data. 
Early research focused on exact forgetting~\cite{ullah2021machine,bourtoule2021machine}, where the model after forgetting is mathematically equivalent to a model retrained on the remaining data. However, these methods require high computational complexity. The initial algorithms were primarily designed for traditional models based on convex functions, which have more uniform structures and make it easier to trace the impact of data~\cite{chen2022graph,wang2023inductive}.

Approximate unlearning reduces the high cost of retraining an entire model by employing data-driven or model-driven approaches. The key to data-driven approximation is achieving forgetting of specific data through data isolation or modification. However, these methods often rely on data partitioning~\cite{neel2021descent} and retraining sub-models~\cite{graves2021amnesiac}, which inherit the substantial computational costs associated with retraining, making them impractical for large-scale, dynamic systems.
In contrast, model-driven approximation~~\cite{giordano2019swiss,koh2017understanding,martens2020new} directly manipulates model parameters, such that the forgotten model becomes indistinguishable from the retrained one in parameter space. Techniques used for model-driven approximation include influence functions~\cite{giordano2019swiss,koh2017understanding}, Fisher information matrix~\cite{martens2020new}, knowledge distillation~\cite{chundawat2023can,chundawat2023zero,kurmanji2023towards}, and stochastic gradient descent~\cite{graves2021amnesiac,wu2020deltagrad}.

Retain-free methods operate only on the original model and the data to be forgotten, without accessing the remaining data. However, existing retain-free methods suffer from flawed assumptions or disruptive mechanisms.
\textit{Gradient ascent}~\cite{graves2021amnesiac,wu2020deltagrad,kurmanji2023towards} disrupts model memory by maximizing loss on forget data, but indiscriminately corrupts shared knowledge, leading to severe performance collapse. 
\textit{Representation misdirection}~\cite{huu2024effects} perturbs features in random directions, yet lacks semantic coherence, resulting in chaotic and unstable representations. 
\textit{Boundary expansion}~\cite{chen2023boundary} introduces artificial classes to shift decision boundaries, but distorts the original output space with unnatural concepts.

In contrast, we propose \textbf{Probing then Editing (PTE)}: a structured, objective-driven unlearning framework that edits knowledge precisely—without global disruption, random guidance, or architectural distortion. PTE conceptualizes unlearning as a ``probe-edit" process. 

\textbf{Probing phase:} The decision boundary neighborhood of the model on the to-be-forgotten class is probed via gradient ascent, and editing instructions are generated based on the model’s own predictions. 

\textbf{Editing phase:} A push-pull collaborative optimization is designed to execute the editing. The ``push" branch actively dismantles the decision region of the forgotten class using the editing instructions, weakening its discriminability. The ``pull" branch employs masked knowledge distillation to anchor the model’s knowledge on retained classes to their original states, ensuring stable model utility. These two branches work collaboratively, enabling efficient and balanced knowledge editing using only the forgotten data and the original model. Our main contributions are as follows:

\begin{itemize}
    \item We propose a novel ``probe-edit" unlearning paradigm, redefining unlearning as a two-stage, goal-directed knowledge editing process.
    \item We design the PTE framework, whose push-pull mechanism enables efficient and balanced knowledge editing without requiring retain data.
    \item We validate the superiority of PTE on multiple industrial benchmarks, demonstrating its outstanding performance in balancing unlearning effectiveness and model utility.
\end{itemize}

\section{Related Work}
\label{sec:related}

\subsection{Machine Unlearning in IIoT}
In IIoT systems, machine unlearning is essential for maintaining model relevance and compliance. Applications include forgetting obsolete failure modes in predictive maintenance~\cite{10102331,10899876,chen2025faultgpt}, removing mislabeled defects in visual inspection~\cite{gu2024anomalygpt}, and updating outdated representations in digital twins~\cite{ren2024industrial}. However, state-of-the-art methods such as Bad Teacher~\cite{chundawat2023can} and SCRUB~\cite{kurmanji2023towards} rely on access to retained data for utility preservation, while Fisher-based~\cite{golatkar2020forgetting,golatkar2020eternal,golatkar2021mixed} and regularization-based~\cite{kurmanji2023towards,tarun2023fast} approaches similarly require original or retained data. This dependency is impractical in IIoT due to data silos and privacy constraints, necessitating retain-free unlearning solutions.

\subsection{Retain-Free Unlearning Methods}
Current retain-free unlearning methods can be broadly categorized into four classes. 
(1) Gradient ascent-based methods~\cite{graves2021amnesiac,wu2020deltagrad,kurmanji2023towards} maximize the loss on forget data to erase memory, but their indiscriminate nature severely damages shared knowledge, leading to catastrophic forgetting. 
(2) Representation misdirection methods~\cite{huu2024effects} apply random or heuristic perturbations to obscure the representation of forget data; however, they lack semantically coherent guidance, resulting in chaotic and unstable editing signals. 
(3) Decision boundary adjustment methods (e.g., Boundary Expand/Shrink~\cite{chen2023boundary}) compress the decision space of the target class by introducing artificial classes, which distorts the model's original output architecture and introduces unnatural biases. 
(4) Model editing methods~\cite{wu2023depn} aim to directly modify model knowledge without retraining. For instance, Influence Unlearn~\cite{giordano2019swiss,koh2017understanding} locates critical parameters via influence functions, but its approximations may fail in complex models; ISPF~\cite{zhang2025toward} leverages generative models to synthesize data for knowledge distillation—though it avoids real retain data, its performance is constrained by the quality of generated samples and involves complex multi-stage training.

\section{Machine Unlearning Formulation}
\label{sec:formulation}

Modern IIoT systems demand dynamic model adaptation where \textbf{class-level unlearning} becomes imperative---particularly when entire operational states (e.g., obsolete failure patterns, deprecated manufacturing modes, or legacy system conditions) must be eradicated from learned representations. We formalize this specialized unlearning paradigm as follows.

\textbf{Class-Centric Problem Setting.}  
We consider a supervised dataset $ \mathcal{D} = \{(x_i, y_i)\}_{i=1}^n $, where:

- Each input $ x_i \in \mathbb{R}^d $ represents a feature vector, such as vibration signals, thermal images, or acoustic patterns.
    
- Each label $ y_i \in \mathcal{Y} = \{1, 2, \dots, K\} $ indicates the class identity corresponding to distinct system states or categories of interest.

Let $ \mathcal{C}_f \subset \{1,\dots,K\} $ denote the set of forget classes requiring erasure (e.g., discontinued failure modes). The dataset is partitioned into a forget set $ \mathcal{D}_f = \{(x_i, y_i) \in \mathcal{D} \mid y_i \in \mathcal{C}_f\} $ and a retained set $ \mathcal{D}_r = \mathcal{D} \setminus \mathcal{D}_f $.

\textbf{Definition 1 (Class-Level Machine Unlearning).}  
Let $ f_{\mathbf{w}_0} $ be a pre-trained model with parameters $ \mathbf{w}_0 $ obtained by minimizing the regularized empirical risk over a hypothesis space $ \mathcal{H} $:
\begin{equation}
    \mathbf{w}_0 = \arg\min_{\mathbf{w} \in \mathcal{H}} \left[ \mathbb{E}_{(x,y)\sim\mathcal{D}} \ell_{\text{CE}}(y, f_\mathbf{w}(x)) + \lambda_{\text{reg}} R(\mathbf{w}) \right],
\end{equation}
where $ f_\mathbf{w}(x) $ is the model's output, $ \ell_{\text{CE}} $ is the cross-entropy loss, $ R(\mathbf{w}) $ is a regularizer (e.g., weight decay), and $ \lambda_{\text{reg}} $ is its coefficient. The unlearning objective is to derive updated model parameters $ \mathbf{w}^- $ from an unlearning algorithm $ \mathcal{U} $, such that:
\begin{equation}
\begin{aligned}
    \mathbf{w}^- &= \mathcal{U}(\mathbf{w}_0, \mathcal{D}_f) \\
    \text{s.t.} \quad
    & \mathbb{E}_{x \sim \mathcal{D}_r} \left[ \text{KL}\left( f_{\mathbf{w}_0}(x) \,\|\, f_{\mathbf{w}^-}(x) \right) \right] \leq \epsilon_{\text{dist}}, \\
    & \text{Conf}_{\mathbf{w}^-}(\mathcal{D}_f) \approx \frac{1}{|\mathcal{Y} \setminus \mathcal{C}_f|},
\end{aligned}
\end{equation}
where $ \epsilon_{\text{dist}} $ is a small tolerance threshold. The first constraint ensures output consistency on retained data using KL-divergence as the distance metric. The second constraint encourages the average confidence of the model on forgotten data to approximate the uniform distribution over the retained classes, implying the model behaves as if it had never seen $ \mathcal{D}_f $. We assume $ |\mathcal{C}_f| = 1 $ for simplicity, denoting the forget class as $ u $.

\section{Methodology}
\label{sec:method}

Directly solving the optimization problem defined in Definition 1---that is, perfectly replicating a retrained model---is computationally prohibitive. Therefore, we propose a more efficient strategy. Our core insight is that effective class-level unlearning can be achieved by structurally editing the model's knowledge through a two-stage ``probe-edit" process, rather than relying on global retraining or access to retain data. This approach, which we call Probing then Editing (PTE), first \textit{probes} the decision boundary of the forget class $\mathcal{C}_f$ to generate targeted \textit{edit instructions}, and then executes a \textit{push-pull} optimization to apply these edits while preserving knowledge on retained classes.

\subsection{Probing the Decision Boundary: Generating Edit Instructions}

To initiate the forgetting process, we probe the model’s sensitivity around the decision boundary of the forget class $\mathcal{C}_f$ using only the forget data $\mathcal{D}_f$ and the original model $f_{\mathbf{w}_0}$. The goal is to generate meaningful \textit{edit instructions} that guide the model to dismantle its knowledge of $\mathcal{C}_f$ in a structured way. This probing phase consists of two steps: noise matrix generation and self-consistent relabeling.

For each sample $(x, y) \in \mathcal{D}_f$, we compute a noise matrix $\delta^*$ that maximizes the model's loss on the original label, effectively pushing the input toward a region where the model becomes uncertain or misclassifies it:
\begin{equation}
    \delta^* = \arg\max_{\|\delta\|_\infty \leq \epsilon} \mathcal{L}(f_{\mathbf{w}_0}(x + \delta), y),
\end{equation}
where $\mathcal{L}$ denotes the cross-entropy loss. We solve this via projected gradient ascent (PGA):
\begin{equation}
    \delta_{t+1} = \Pi_{\epsilon} \left( \delta_t + \eta \cdot \nabla_\delta \mathcal{L}(f_{\mathbf{w}_0}(x + \delta_t), y) \right),
\end{equation}
where $\Pi_{\epsilon}$ denotes the projection onto the $\epsilon$-ball (e.g., clipping to $[-\epsilon, \epsilon]$ under the $L_\infty$ norm), and $\eta$ is the step size.

The resulting probed input $x_{\text{probe}} = x + \delta^*$ resides near the decision boundary of $\mathcal{C}_f$, where the model’s prediction is most malleable. We then generate an \textit{edit instruction} by assigning a new label $y_{\text{edit}}$ based on the model’s own prediction at this probed point:
\begin{equation}
    y_{\text{edit}} = \arg\max f_{\mathbf{w}_0}(x_{\text{probe}}), \quad \text{with } y_{\text{edit}} \neq y.
\end{equation}
This \textit{self-consistent supervision} ensures that the edit signal is coherent with the model’s current knowledge state, avoiding arbitrary or disruptive labels. The pair $(x_{\text{probe}}, y_{\text{edit}})$ forms a synthesized instruction that tells the model: ``This input, which you previously associated with class $y$, should now be strongly associated with class $y_{\text{edit}}$." The full set of such instructions is denoted as $\mathcal{D}_E = \{(x_{\text{probe}}, y_{\text{edit}})\}$.

The noise matrix $\delta^*$ has the same size as the input $x$. For example, if the input is an image, the shape of $\delta^*$ matches the image dimensions (e.g., $H \times W \times C$, where $H$, $W$, and $C$ represent height, width, and channels, respectively). This design enables the probe to precisely manipulate local features or pixels, allowing fine-grained exploration of the model's decision boundary for class $\mathcal{C}_f$.

\paragraph{One-shot Probing}
This probing phase is performed in a \textit{one-shot} manner: all edit instructions are generated in a single forward pass over $\mathcal{D}_f$, with the model parameters $\mathbf{w}_0$ kept frozen. This ensures high efficiency and scalability, making it suitable for both single-class and multi-class unlearning scenarios.

\paragraph{Multi-Class Unlearning}

If multiple classes need to be forgotten simultaneously, we can learn a separate noise matrix for each class. Since the optimization process only involves the noise matrices themselves, and the model weights remain frozen throughout, the computational overhead is almost negligible. Whether for single-class or multi-class unlearning, the overall algorithm needs to be executed only once. This method is not only efficient but also flexible enough to handle different scales of unlearning tasks.

\begin{algorithm}[tb]
\caption{Probing then Editing (PTE)}
\label{alg:pte}
\SetAlgoLined
\DontPrintSemicolon
\KwIn{Pre-trained model $f_{\mathbf{w}_0}$, Forget set $\mathcal{D}_f$, Probing radius $\epsilon$, PGA iterations $T$, Learning rate $\eta$, Distillation temperature $T$, Fine-tuning epochs $E$, Push/Pull learning rates $\eta_{\text{push}}, \eta_{\text{pull}}$}
\KwOut{Unlearned model $f_{\mathbf{w}^-}$}

\textbf{Phase 1: One-shot Probing} \\
$\mathcal{D}_E \leftarrow \emptyset$ \\
\For{each $(X, Y) \in \mathcal{D}_f$ (mini-batch)}{
    $\delta \sim \mathcal{N}(0, I)$; \ $\delta \leftarrow \text{clip}(\delta, -\epsilon, \epsilon)$ \\
    \For{$t = 0$ \KwTo $T-1$}{
        $\delta \leftarrow \Pi_{\epsilon} \left( \delta + \eta \cdot \nabla_{\delta} \mathcal{L}(f_{\mathbf{w}_0}(X + \delta), Y) \right)$
    }
    $X_{\text{probe}} \leftarrow X + \delta$; \ 
    $Y_{\text{edit}} \leftarrow \arg\max f_{\mathbf{w}_0}(X_{\text{probe}})$ \\
    \For{each $(x_{\text{probe}}, y_{\text{edit}}, y) \in (X_{\text{probe}}, Y_{\text{edit}}, Y)$}{
        \If{$y_{\text{edit}} \neq y$}{
            Add $(x_{\text{probe}}, y_{\text{edit}})$ to $\mathcal{D}_E$
        }
    }
}

\textbf{Phase 2: Push-Pull Knowledge Editing} \\
\For{$e = 1$ \KwTo $E$}{
    Sample $(x_{\text{probe}}, y_{\text{edit}}) \sim \mathcal{D}_E$ \\
    $\mathbf{w} \leftarrow \mathbf{w} - \eta_{\text{push}} \nabla_{\mathbf{w}} \mathcal{L}(f_{\mathbf{w}}(x_{\text{probe}}), y_{\text{edit}})$ \\
    
    Sample $(x, y) \sim \mathcal{D}_f$ \\
    $\tilde{\mathbf{p}} \leftarrow \operatorname{Normalize}\left( \operatorname{Mask}_u\left( \operatorname{Softmax}(f_{\mathbf{w}_0}(x)/T) \right) \right)$ \\
    $\mathcal{L}_{\text{pull}} \leftarrow T^2 \cdot \mathrm{KL}\left( \tilde{\mathbf{p}} ~\|~ \operatorname{Softmax}(f_{\mathbf{w}}(x)/T) \right)$ \\
    $\mathbf{w} \leftarrow \mathbf{w} - \eta_{\text{pull}} \nabla_{\mathbf{w}} \mathcal{L}_{\text{pull}}$
}
\Return $f_{\mathbf{w}^-}$
\end{algorithm}

\subsection{Push-Pull Knowledge Editing}
Armed with the edit instructions $\mathcal{D}_E$, we perform a dual-phase fine-tuning process to edit the model’s knowledge. Crucially, this stage operates without any access to retain data, relying only on $\mathcal{D}_f$ and the original model $f_{\mathbf{w}_0}$. The optimization follows a \textit{push-pull} mechanism: the push branch actively dismantles the decision structure of the forget class, while the pull branch passively anchors the model’s behavior on non-forget classes to their original state.

\paragraph{Push:}
The push branch uses the edit instructions $\mathcal{D}_E$ as training data to force the model to reconfigure its decision boundary for $\mathcal{C}_f$. We minimize the cross-entropy loss with a relatively high learning rate $\eta_{\text{push}}$:
\begin{equation}
\label{eq:push}
    \mathcal{L}_{\text{push}} = \mathbb{E}_{(x_{\text{probe}}, y_{\text{edit}}) \sim \mathcal{D}_E} \left[ \ell_{\text{CE}}(f_\mathbf{w}(x_{\text{probe}}), y_{\text{edit}}) \right].
\end{equation}
By training the model to confidently predict $y_{\text{edit}}$ on inputs originally belonging to $\mathcal{C}_f$, this process actively ``pushes'' the decision region of $\mathcal{C}_f$ toward collapse, effectively erasing its discriminative capability.

\paragraph{Pull:}
To preserve model utility on retained classes without accessing retain data, we employ masked knowledge distillation from the original model $f_{\mathbf{w}_0}$. For each $x \in \mathcal{D}_f$, we compute the teacher logits $\mathbf{z} = f_{\mathbf{w}_0}(x)$ and apply the softmax function with a temperature parameter $T > 0$ to obtain a softened probability distribution:
\begin{equation}
    \mathbf{p} = \operatorname{Softmax}(\mathbf{z}/T).
\end{equation}
We then apply a masking operation $\operatorname{Mask}_u(\cdot)$ that suppresses the probability of the forget class $u$ by setting it to zero, while preserving the values for all other classes:
\begin{equation}
    \operatorname{Mask}_u(\mathbf{p})_k = \begin{cases}
      p_k & \text{if } k \neq u \\
      0   & \text{if } k = u
   \end{cases}.
\end{equation}
To ensure the resulting vector remains a valid probability distribution, we apply a normalization step:
\begin{equation}
    \tilde{\mathbf{p}} = \operatorname{Normalize}\left( \operatorname{Mask}_u\left( \operatorname{Softmax}(\mathbf{z}/T) \right) \right),
\end{equation}
where $\operatorname{Normalize}(\mathbf{v}) = \mathbf{v} / \sum_k v_k$ scales the vector such that its entries sum to one. This construction ensures that: (i) the output probability for the forget class is driven to zero (forgetting condition), (ii) the relative magnitudes among retained classes are preserved according to the original model’s knowledge, and (iii) $\tilde{\mathbf{p}}$ is a valid categorical distribution (probability condition).

The student model $f_{\mathbf{w}}$ is trained to match this masked and normalized target distribution using temperature-scaled knowledge distillation. Specifically, we minimize the following KL divergence loss:
\begin{equation}
\label{eq:pull}
    \mathcal{L}_{\text{pull}} = T^2 \cdot \mathbb{E}_{x \sim \mathcal{D}_f} \left[ \operatorname{KL}\left( \operatorname{Softmax}(f_\mathbf{w}(x)/T) \,\|\, \tilde{\mathbf{p}} \right) \right].
\end{equation}
The $T^2$ scaling factor ensures that the gradient magnitude remains consistent when using high temperatures. 

This ``pull'' objective anchors the model’s predictions on non-forget classes to those of the original model, thereby stabilizing predictive behavior across retained knowledge and mitigating performance degradation during unlearning.

\paragraph{Training Procedure.}
The final unlearning algorithm is summarized in Algorithm~\ref{alg:pte}. We first generate the edit instructions $\mathcal{D}_E$ (Phase 1). Then, we perform alternating optimization (Phase 2): for each epoch, we first execute a push step using $\mathcal{D}_E$, followed by a pull step using $\mathcal{D}_f$. This entire process requires only the forget set $\mathcal{D}_f$ and the original model $f_{\mathbf{w}_0}$, making it a truly retain-free solution.

\section{Experiments}
\label{sec:experiments}

To align with practical concerns such as privacy preservation, restricted data access, and limited computational resources, we enforce two key constraints in our experiments:
(i) the model is not allowed to access any retained data;
(ii) no intervention is permitted during pre-training — all models are trained from scratch.
Further implementation details are provided in the supplementary material.

\subsection{Experimental Setup}

\begin{table*}[t]
    \centering
    \caption{Performance comparison of single-class forgetting across different unlearning methods on CIFAR-10 and CIFAR-100 datasets. 
    \g{Gray} indicates methods with retain data or intervention, and \textbf{bold} indicates the single best result among methods w/o retain data or intervention (if multiple, results are not in bold). The same notation applies hereafter.
    }
    \resizebox{1.7\columnwidth}{!}{
        \begin{tabular}{cc|cccccc|cccccc}
            \toprule
            \multicolumn{2}{c|}{\multirow{2}{*}{Method}} &
            \multicolumn{6}{c|}{CIFAR-10} & \multicolumn{6}{c}{CIFAR-100} \\
            \cmidrule(lr){3-8} \cmidrule(lr){9-14}
                &
                & $\mathrm{Acc}_{\mathrm{f}} \downarrow$ 
                & $\mathrm{Acc}_{\mathrm{r}} \uparrow$ 
                & $\mathrm{Acc}_{\mathrm{ft}} \downarrow$ 
                & $\mathrm{Acc}_{\mathrm{rt}} \uparrow$ 
                & $\mathrm{H\text{-}Mean} \uparrow$ 
                & $\mathrm{MIA} \downarrow$ 
                & $\mathrm{Acc}_{\mathrm{f}} \downarrow$ 
                & $\mathrm{Acc}_{\mathrm{r}} \uparrow$ 
                & $\mathrm{Acc}_{\mathrm{ft}} \downarrow$ 
                & $\mathrm{Acc}_{\mathrm{rt}} \uparrow$ 
                & $\mathrm{H\text{-}Mean} \uparrow$ 
                & $\mathrm{MIA} \downarrow$ \\
        \midrule
            \multicolumn{2}{c|}{Original Model} & 99.40 & 99.10 & 96.50 & 93.69 & - & 98.00 & 98.80 & 97.33 & 73.00 & 77.10 & - & 98.44 \\
            \multicolumn{2}{c|}{Retrain Model} & 0 & 99.71 & 0 & 95.00 & 97.10 & - & 0 & 94.87 & 0 & 76.65 & 75.09 & -\\
        \midrule
            \multicolumn{2}{c|}{Random Label} &17.12 & 95.88 &35.50 &90.62 & 77.03 & 0.09 & 6.20 & 90.90 & 3.00 & 70.66 & 70.33 & 0.03\\
            \multicolumn{2}{c|}{Negative Gradient} & 0 & 17.55 &0 & 17.63 & 29.79 & 8.75 & 1.40 & 89.72 & 1.00 & 69.78 & 70.87 & 0.11\\
            \multicolumn{2}{c|}{Boundary Shrink} &1.08 & 74.76 &3.80 & 74.04 & 83.36 &  0.32 & 5.20 & 90.50 & 2.00 & 70.34 & 70.64 & 0.20\\
            \multicolumn{2}{c|}{Boundary Expand} &20.88 & 96.61 & 28.70 & 91.00 & 81.52 & 0.80 & 3.60 & 91.06 & 2.00 & 70.75 & 70.87 & 0.15\\
            \multicolumn{2}{c|}{DELETE} & 2.62 & 98.82 & 2.90 & 93.48 & 95.12 & 0.01 & 4.60 & 91.40 & 3.00 & 71.23 & 70.61 & 0.02\\
            \multicolumn{2}{c|}{Bad Teacher} &45.92 & 95.41 & 67.40 & 89.96 & 78.62 & 0.27 & 4.80 & 90.88 & 3.00 & 70.46 & 70.23 & 0.22\\
            \multicolumn{2}{c|}{IMP} & 0 & 19.20 & 0 & 19.59 & 35.30 & 1.00 & 2.40 & 90.13 & 1.00 & 70.03 & 70.54 & 0.01\\
            \multicolumn{2}{c|}{\g{SCRUB}} &\g{0.44} & \g{98.09} & \g{0.40} & \g{93.94} & \g{96.68} & \g{0} & \g{0} & \g{96.55} & \g{0} & \g{75.41} & \g{71.75} & \g{0}\\
            \multicolumn{2}{c|}{\g{Finetune}} &\g{98.16} & \g{99.38} & \g{93.30} & \g{94.00} & \g{6.21} & \g{0.91} & \g{97.00} & \g{97.93} & \g{65.00} & \g{77.68} & \g{14.51} & \g{0.85}\\
            \multicolumn{2}{c|}{SURE} & 10.16 & 98.19 & 9.75 & 90.08 & 82.35 & 10.07 & 2.35 & 88.70 & 3.12 & 64.37  & 61.45 & 1.47\\
        \midrule
            \multicolumn{2}{c|}{PTE (Ours)} & 0 & \textbf{99.02} & \textbf{0} & \textbf{94.11} & \textbf{96.81} & 0 & \textbf{0} & \textbf{95.00} & \textbf{0} & \textbf{75.23} & \textbf{74.76} & \textbf{0}\\
            \bottomrule
        \end{tabular}
    }
    \label{tab:cifar10_cifar100}
\end{table*}

\textbf{Datasets.} To ensure fair comparison with prior work, we conduct experiments on CIFAR-10 and CIFAR-100. In addition, we evaluate our method on two Industrial IoT datasets—CWRU and SCUT-FD, as detailed in the following section.

\textbf{Models.} We adopt ResNet-18 as the default backbone in our experiments. Additionally, we evaluate our method on ViT-s~\cite{dosovitskiy2020image} and Swin-T~\cite{liu2021swin} to verify its applicability to transformer-based architectures.

\textbf{Compared Methods.} In our study, we compare several representative machine unlearning baselines, including Random Label, Gradient Ascent, Finetune, Boundary Shrink~\cite{chen2023boundary}, Boundary Expand~\cite{chen2023boundary}, IMP~\cite{cha2024learning}, Bad Teacher~\cite{chundawat2023can}, DELETE~\cite{zhou2025decoupled}, SCRUB~\cite{kurmanji2023towards}, and SURE~\cite{sepahvand2025selective}.

We also report the performance of the original model and the retrained model, the latter of which is commonly regarded as the gold standard in machine unlearning~\cite{thudi2022necessity}.

\textbf{Evaluation Metrics.}  
We assess unlearning effectiveness using four accuracy measures: 
(1) $\text{Acc}_{f}$ on the forget training set $\mathcal{D}_f$, 
(2) $\text{Acc}_{ft}$ on the forget test set $\mathcal{D}_{ft}$, 
(3) $\text{Acc}_{r}$ on the retain training set $\mathcal{D}_r$, and 
(4) $\text{Acc}_{rt}$ on the retain test set $\mathcal{D}_{rt}$. 

To jointly reflect the trade-off between forgetting and retention, we compute the harmonic mean (H-Mean)~\cite{zhao2024continual} of $\text{Acc}_{rt}$ and the performance drop on $\mathcal{D}_{ft}$ after unlearning ($\text{Drop}_{ft}$). 
We also report the success rate of Membership Inference Attack (MIA) as an indicator of privacy leakage. A lower MIA score suggests better forgetting, as the model is less likely to memorize individual samples from the forget set~\cite{fan2023salun}.
Ideally, a good unlearning method should reduce $\text{Acc}_{f}$, $\text{Acc}_{ft}$, and MIA, while maintaining high $\text{Acc}_{rt}$ and H-Mean to preserve performance on retained classes.
For a detailed description of these evaluation metrics, including their formal definitions and computation procedures, we refer readers to the \textbf{supplementary material}.

\subsection{Results and Comparisons}

\subsubsection{Performance on Different Datasets.}

\noindent
We evaluate single-class forgetting performance across different datasets. Although our setting prohibits access to retain data, we include two representative baselines—Finetune and SCRUB—that leverage such privileges, highlighted in gray for reference.

Using CIFAR-10 as an example (Table~\ref{tab:cifar10_cifar100}), all methods except Finetune demonstrate varying degrees of forgetting. DELETE reduces both $\mathrm{Acc}_{\mathrm{f}}$ and $\mathrm{Acc}_{\mathrm{ft}}$ to around 3\%, indicating incomplete forgetting. Negative Gradient drives $\mathrm{Acc}_{\mathrm{f}}$ and $\mathrm{Acc}_{\mathrm{ft}}$ to 0\%, but severely degrades $\mathrm{Acc}_{\mathrm{r}}$ and $\mathrm{Acc}_{\mathrm{rt}}$ (only 20\%). In contrast, our method PTE achieves complete forgetting in terms of $\mathrm{Acc}_{\mathrm{ft}}$, while maintaining $\mathrm{Acc}_{\mathrm{rt}}$ nearly on par with the retrained model (94.11\% vs. 95.00\%). Notably, PTE yields a near-optimal H-Mean (96.81 vs. 97.10) and 0 MIA. When extended to the larger CIFAR-100 dataset, PTE continues to outperform all other baselines.

\subsubsection{Performance on Different Models.}

We conduct single-class forgetting experiments on Swin-T and ViT-S, with results on CIFAR-10 reported in Table~\ref{tab:model}. 
On Swin-T, DELETE achieves high retained accuracy ($\mathrm{Acc}_{\mathrm{r}}$ and $\mathrm{Acc}_{\mathrm{rt}}$), but exhibits a non-negligible forgetting gap with $\mathrm{Acc}_{\mathrm{f}} = 1.06$. 
In contrast, PTE achieves perfect forgetting ($\mathrm{Acc}_{\mathrm{f}} = 0$) while maintaining retention performance within $0.4$/$0.44\%$ of full retraining, demonstrating retraining-level utility preservation. 
On ViT-S, PTE further improves $\mathrm{Acc}_{\mathrm{r}}$ to $71.65$ ($+0.99\%$) and $\mathrm{Acc}_{\mathrm{rt}}$ to $69.05$ ($+1.51\%$), outperforming both retraining and DELETE in model retention. 
Collectively, PTE achieves complete forgetting and superior retention across two dominant vision Transformer architectures, yielding the most balanced performance among current methods.

\begin{table}[t]
    \centering
    \caption{Performance comparison across different models.}
    \resizebox{0.8\columnwidth}{!}{
        \begin{tabular}{cc|cccc}
            \toprule
                \multicolumn{2}{c}{}  
                & $\mathrm{Acc}_{\mathrm{f}} \downarrow$ 
                & $\mathrm{Acc}_{\mathrm{r}} \uparrow$ 
                & $\mathrm{Acc}_{\mathrm{ft}} \downarrow$ 
                & $\mathrm{Acc}_{\mathrm{rt}} \uparrow$ \\
            \midrule
            \multicolumn{6}{l}{\emph{Swin-T}\vspace{0.02in}}\\
            \multicolumn{2}{l}{ Original Model} & 81.48 & 82.62 & 77.80 & 79.91  \\
            \multicolumn{2}{l}{ Retrain Model} & 0 & 83.37 & 0 & 80.87  \\
            \multicolumn{2}{l}{ DELETE} & 1.06 & \textbf{83.78} & 0 & \textbf{81.22}  \\
            \multicolumn{2}{l}{ Bad Teacher} & 8.72 & 47.24 & 9.50 & 38.79  \\
            \multicolumn{2}{l}{ PTE (Ours)} & \textbf{0} & 82.97 & 0 & 80.43  \\
            \midrule
            \multicolumn{6}{l}{\emph{ViT-S}\vspace{0.02in}}\\
            \multicolumn{2}{l}{ Original Model} & 65.02 & 69.80 & 61.70 & 67.36 \\
            \multicolumn{2}{l}{ Retrain Model} & 0 & 70.66 & 0 & 67.54  \\
            \multicolumn{2}{l}{ DELETE} & 0.12 & 71.32 & 0 & 68.18  \\
            \multicolumn{2}{l}{ Bad Teacher} & 9.04 & 50.70 & 11.30 & 52.50  \\
            \multicolumn{2}{l}{ PTE (Ours)} & \textbf{0} & \textbf{71.65} & 0 & \textbf{69.05}  \\
            \bottomrule
        \end{tabular}
    }
    \label{tab:model}
\end{table}

\subsection{Stability Analysis}  

\begin{figure}[htbp]
    \centering
    \includegraphics[width = 1.0\linewidth]{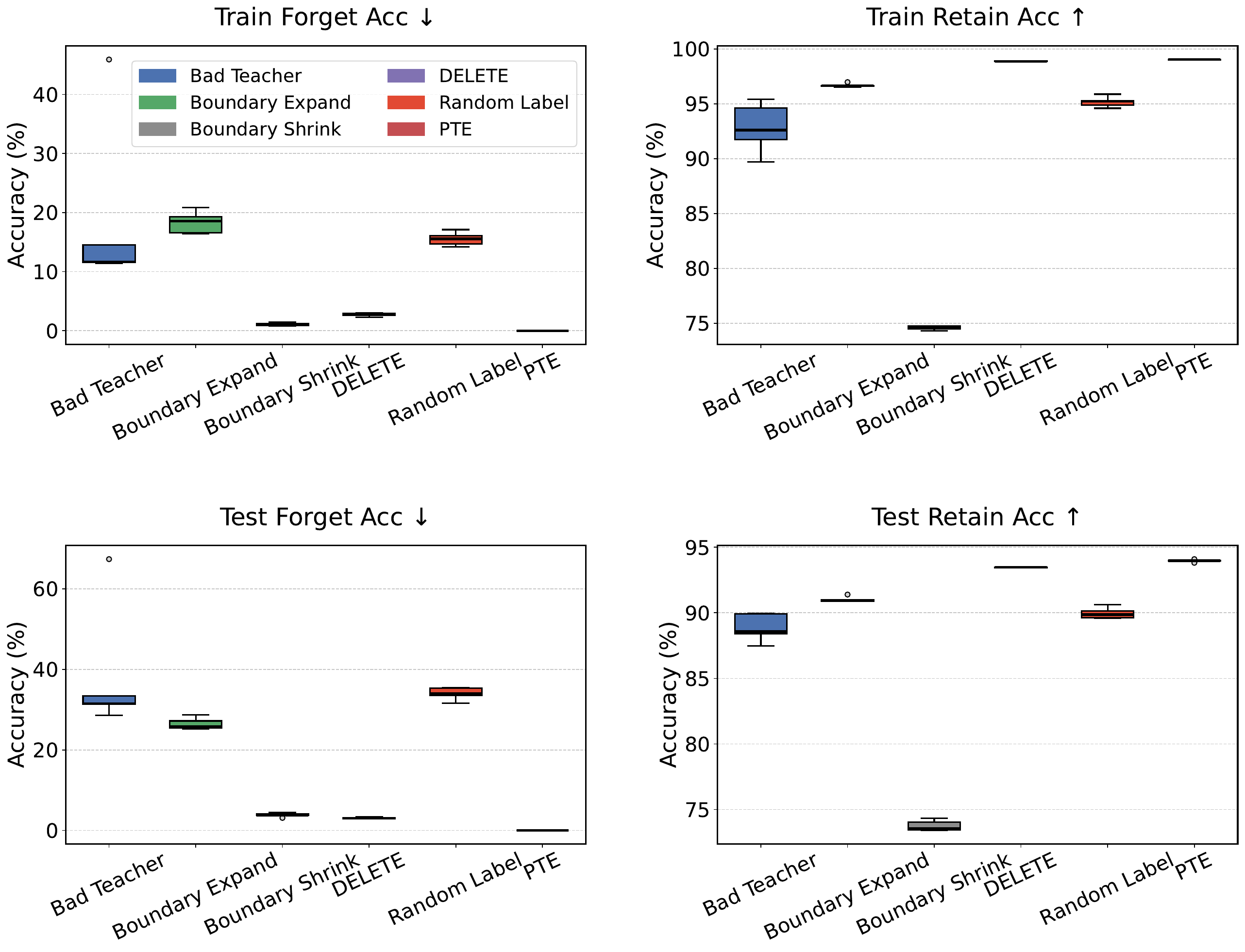}
    \caption{Boxplot comparison.}
    \label{fig:boxplot}
\end{figure}

To assess the robustness of each approach, we performed five repeated single‑class forgetting experiments on CIFAR-10. As shown in Figure~\ref{fig:boxplot}, some methods (e.g., Bad Teacher and Boundary Expand) display considerable performance variability, reflected in the large widths of their box plots. By contrast, our method exhibits excellent stability.

\begin{figure}[htbp]
    \centering
    \includegraphics[width = 0.95\linewidth]{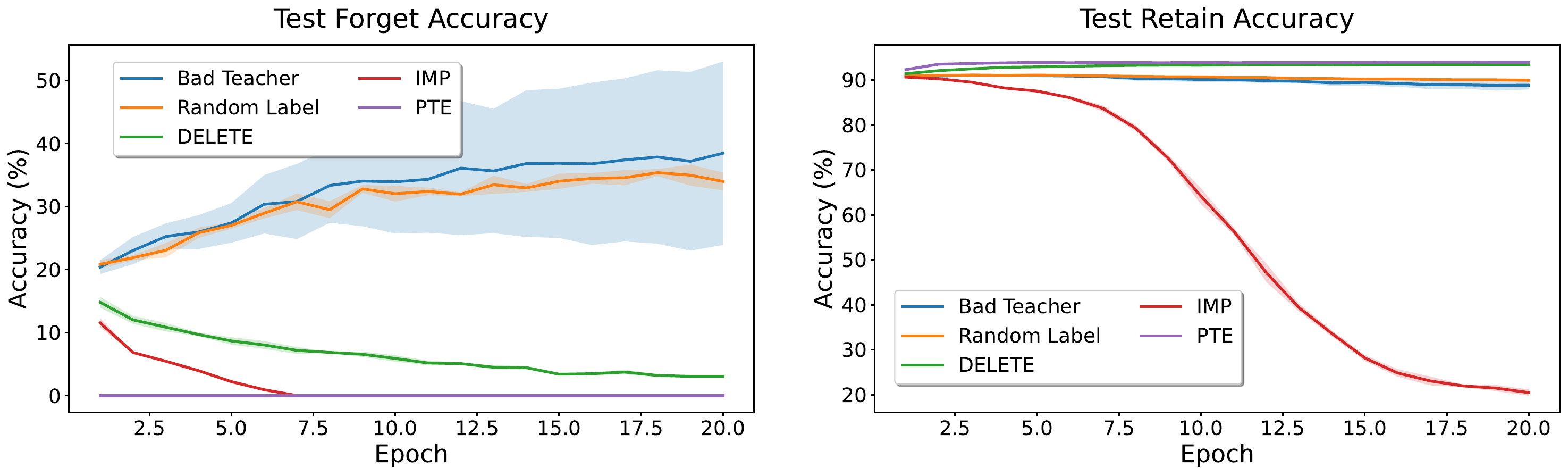}
    \caption{Epoch‑wise Test Forget and Retain Accuracy.}
    \label{fig:test_forget_retain}
\end{figure}

Figure~\ref{fig:test_forget_retain} shows the 20‑epoch learning trajectories for the five repeated single‑class forgetting experiments. Bad Teacher, Random Label, and IMP exhibit substantial performance fluctuations. Although DELETE demonstrates outstanding stability, our PTE achieves a faster decline in $\mathrm{Acc}_{\mathrm{ft}}$ while also surpassing DELETE in $\mathrm{Acc}_{\mathrm{rt}}$.

\subsection{Multi-Class Unlearning}

We conduct multi‑class forgetting experiments on CIFAR‑100, as reported in Table~\ref{tab:multi_class}. The results indicate that, as the number of forgotten classes increases, the performance of several baselines deteriorates markedly. When five classes are removed, the $\mathrm{Acc}_{\mathrm{ft}}$ of all other methods exceeds 7.5\%, in sharp contrast to their behavior in the single‑class setting. By contrast, our approach exhibits outstanding multi‑class forgetting capability, underscoring the strength of the proposed \emph{Push–Pull Knowledge Editing} paradigm.

\begin{table}[t]
    \centering
    \caption{Multi-class Unlearning Performance on CIFAR-100.}
    \resizebox{\columnwidth}{!}{
    \begin{tabular}{cc|cc|cc|cc}
        \toprule
        \multicolumn{2}{c|}{\multirow{2}{*}{Method}} & 
        \multicolumn{2}{c|}{1 Class} & 
        \multicolumn{2}{c|}{5 Classes} & 
        \multicolumn{2}{c}{10 Classes} \\
        \cmidrule(lr){3-4} \cmidrule(lr){5-6} \cmidrule(lr){7-8}
          & 
            & $\mathrm{Acc}_{\mathrm{ft}} \downarrow$ 
            & $\mathrm{Acc}_{\mathrm{rt}} \uparrow$ 
            & $\mathrm{Acc}_{\mathrm{ft}} \downarrow$ 
            & $\mathrm{Acc}_{\mathrm{rt}} \uparrow$ 
            & $\mathrm{Acc}_{\mathrm{ft}} \downarrow$ 
            & $\mathrm{Acc}_{\mathrm{rt}} \uparrow$ \\
        \midrule 
        \multicolumn{2}{c|}{Retrain Model}      & 0 & 76.65 & 0 & 77.31 & 0 & 77.58 \\
        \multicolumn{2}{c|}{DELETE}    & 3.00 & 71.23 & 8.40 & 75.09 & 11.90 & 76.10 \\
        \multicolumn{2}{c|}{Boundary Shrink}    & 2.00 & 70.34 & 7.80 & 68.58 & 3.20 & 53.76 \\
        \multicolumn{2}{c|}{Bad Teacher}        & 3.00 & 70.46 & 9.40 & 69.05 & 5.30 & 60.87 \\
        \midrule
        \multicolumn{2}{c|}{PTE (Ours)} & \textbf{0} & \textbf{75.23} & \textbf{0} & \textbf{77.18} & \textbf{0} & \textbf{77.53} \\
        \bottomrule
    \end{tabular}    
    \label{tab:multi_class}
    }
\end{table}

\begin{table}[t]
    \centering
    \caption{Ablation study.}
    \resizebox{0.8\columnwidth}{!}{
        \begin{tabular}{cccccc}
            \toprule
                \multicolumn{2}{c}{}  
                & $\mathrm{Acc}_{\mathrm{f}}\!\!\downarrow$ 
                & $\mathrm{Acc}_{\mathrm{r}}\!\!\uparrow$ 
                & $\mathrm{Acc}_{\mathrm{ft}}\!\!\downarrow$ 
                & $\mathrm{Acc}_{\mathrm{rt}}\!\!\uparrow$ \\
            \midrule
            \multicolumn{2}{c}{Retrain Model}              & 0 & 94.87 & 0 & 76.65  \\
            \multicolumn{2}{c}{PTE (Ours)}  & \textbf{0} & \textbf{95.00} & \textbf{0} & \textbf{75.23}  \\
            \midrule
            \multicolumn{5}{l}{\emph{Core component ablation}} \\[0.4ex]
            \multicolumn{2}{l}{ w/o Pull (Eq.~\eqref{eq:pull})}           & 15.60 & 84.61 & 13.00 & 66.79  \\
            \multicolumn{2}{l}{ w/o Push (Eq.~\eqref{eq:push})}           & 7.85 & 92.77 & 5.16 & 72.25 \\
            \midrule
            \multicolumn{5}{l}{\emph{Scheduling ablation}} \\[0.4ex]
            \multicolumn{2}{l}{ Push then Pull} & 2.60 & 76.19 & 3.00 & 59.82  \\
            \multicolumn{2}{l}{ Pull then Push} & 14.22 & 86.52 & 10.91 & 67.31  \\
            \bottomrule
        \end{tabular}
    }
    \label{tab:ablation}
\end{table}

\begin{figure}[htbp]
    \centering
    \includegraphics[width = 0.9\linewidth]{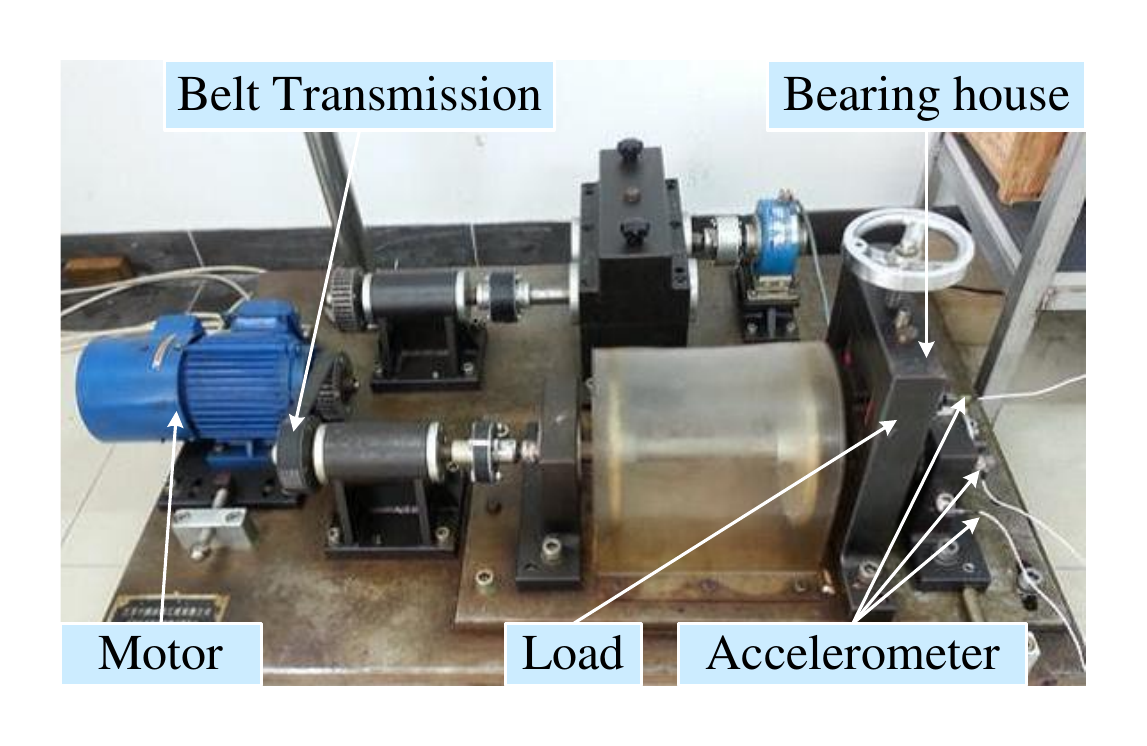}
    \caption{Our vibration signal acquisition equipment.}
    \label{fig:scut_fd}
\end{figure}

\subsection{Ablation Study}

We perform systematic ablation studies on the CIFAR-100 dataset, focusing on the role of the Push-Pull Knowledge editing. As shown in Table~\ref{tab:ablation}, removing either the Push or Pull module leads to a notable performance drop. Specifically, removing the Pull operation results in incomplete forgetting ($\mathrm{Acc}_{\mathrm{f}}$) and degraded retention performance ($\mathrm{Acc}_{\mathrm{r}}$), indicating that the Pull step plays a critical role in identifying and suppressing the activation paths of the target class. In contrast, removing the Push operation impairs the model’s ability to uniformly erase the target knowledge. These findings confirm that both operations are indispensable and complementary for achieving selective and effective unlearning.

We further compare different scheduling strategies to demonstrate the superiority of our alternating Push-Pull optimization. On CIFAR-100, naive sequential schedules, such as ``Push then Pull'' or ``Pull then Push'' yield suboptimal trade-offs, either failing to remove the target knowledge or harming the retention of non-target classes. In contrast, our alternating Push-Pull scheme gradually adjusts the model’s decision boundaries in both directions. It achieves a more stable trade-off between forgetting and retention, and approaches the performance of full retraining while requiring no access to retained data.

\section{Application to Industrial IoT Tasks}
\label{sec:app}

\begin{table}[ht]
\centering
\caption{Summary of datasets used in experiments.}
\label{tab:cwru_scutfd}
\resizebox{\linewidth}{!}{
\begin{tabular}{@{}lcc@{}}
\toprule
\textbf{} & \textbf{CWRU} & \textbf{SCUT-FD} \\ \midrule
\textbf{Input Shape (C$\times$L)} & 2$\times$1024 & 9$\times$1024 \\
\textbf{Train / Test Size} & 22,400 / 5,600 & 12,600 / 2,800 \\
\textbf{\#Classes} & 10 & 7 \\
\textbf{Forget Class (Index)} & 0 (e.g., Inner Race Fault 0.007) & 6 (e.g., Normal Condition) \\ \bottomrule
\end{tabular}
}
\end{table}

\begin{table*}[t]
    \centering
    \caption{Results on the CWRU and SCUT-FD datasets.}
    \resizebox{1.7\columnwidth}{!}{
        \begin{tabular}{cc|cccccc|cccccc}
            \toprule
            \multicolumn{2}{c|}{\multirow{2}{*}{Method}} &
            \multicolumn{6}{c|}{CWRU} & \multicolumn{6}{c}{SCUT-FD} \\
            \cmidrule(lr){3-8} \cmidrule(lr){9-14}
                &
                & $\mathrm{Acc}_{\mathrm{f}} \downarrow$ 
                & $\mathrm{Acc}_{\mathrm{r}} \uparrow$ 
                & $\mathrm{Acc}_{\mathrm{ft}} \downarrow$ 
                & $\mathrm{Acc}_{\mathrm{rt}} \uparrow$ 
                & $\mathrm{H\text{-}Mean} \uparrow$ 
                & $\mathrm{MIA} \downarrow$ 
                & $\mathrm{Acc}_{\mathrm{f}} \downarrow$ 
                & $\mathrm{Acc}_{\mathrm{r}} \uparrow$ 
                & $\mathrm{Acc}_{\mathrm{ft}} \downarrow$ 
                & $\mathrm{Acc}_{\mathrm{rt}} \uparrow$ 
                & $\mathrm{H\text{-}Mean} \uparrow$ 
                & $\mathrm{MIA} \downarrow$ \\
        \midrule
            \multicolumn{2}{c|}{Original Model} &98.92 & 99.04 & 91.22 & 92.05 & - & 97.95 & 100 & 99.08 &70.12 & 80.79  & - & 98.23 \\
            \multicolumn{2}{c|}{Retrain Model} & 0 & 98.71 & 0 & 92.46 & 95.10 & - &0 & 99.07 & 0 & 80.64 & 73.93 & -\\
        \midrule
            \multicolumn{2}{c|}{Negative Gradient} & 10.15 & 83.65 & 6.28 & 75.25 & 81.40 & 7.55 & 2.30 & 97.08 & 3.03 & 73.55 & 69.72 & 1.00\\
            \multicolumn{2}{c|}{Boundary Shrink} & 6.65 & 91.35 & 6.60 & 83.97 & 83.42 & 2.80 & 1.56 & 96.54 & 1.90 & 71.45 & 68.27 & 0.50 \\
            \multicolumn{2}{c|}{DELETE} & 1.75 & 96.58 & 1.06 & 94.21 & 92.58 & 0.05 & 0.50 & 97.70 & 1.00 & 76.65 & 72.32 & 1.25 \\
            \multicolumn{2}{c|}{Bad Teacher} &1.95 & 92.32 & 17.50 & 88.15 & 82.46 & 0.55 & 1.80 & 97.46 & 3.50 & 73.92 & 68.94 & 0.20 \\
        \midrule
            \multicolumn{2}{c|}{PTE (Ours)} & \textbf{0} & \textbf{98.66} & \textbf{0} & \textbf{92.00} & \textbf{94.78} & 0 & \textbf{0} & \textbf{98.45} & \textbf{0} & \textbf{78.90} & \textbf{73.70} & \textbf{0}\\
            \bottomrule
        \end{tabular}
    }
    \label{tab:cwru_scut-fd}
\end{table*}

In this section, we apply machine unlearning techniques to achieve fault diagnosis of mechanical equipment in IIoT environments. Figure~\ref{fig:scut_fd} shows a rotating machinery system, and the goal of fault diagnosis is to analyze operational data to identify and localize potential faults. We use the CWRU and SCUT-FD datasets for our experiments:

\textbf{CWRU}~\cite{smith2015rolling} is a standard dataset for bearing fault diagnosis in rotating machinery. It contains vibration signals collected under different motor loads and fault conditions, including inner race, outer race, and ball defects. Fault diameters vary from 0.007 inches to 0.021 inches. Each sample is segmented into windows of 1024 points across 2 channels, reflecting fine-grained mechanical degradation.

\textbf{SCUT–FD.}
The SCUT–FD dataset~\cite{9093960,LI2022108487} comprises multi-sensor vibration recordings from nine accelerometers mounted on a purpose-built bearing-fault platform. Data were captured at four shaft speeds (500, 800, 1100 and 1400 rpm) to reflect realistic load variations, covering seven bearing conditions: normal (NC); inner-ring defects of 0.5, 1 and 2 mm diameter (IF-0.5, IF-1, IF-2); and outer-ring defects of the same severities (OF-0.5, OF-1, OF-2).

As summarized in Table~\ref{tab:cwru_scutfd}, one class is selected from each dataset as the \emph{forget set}, while the retain classes form the \emph{retain set}. This setting allows us to evaluate whether the model can selectively forget target concepts without impairing its performance on the rest.

We compare PTE with other common machine unlearning methods, as shown in Table~\ref{tab:cwru_scut-fd}. PTE outperforms all other methods across all evaluation metrics, and its performance is closest to that of the retrained model, demonstrating its effectiveness in fault diagnosis tasks.

\subsection{Feature Representation Visualization}

\begin{figure}[ht]
    \centering
    \begin{minipage}{0.4\linewidth}
        \centering
        \includegraphics[width=\linewidth]{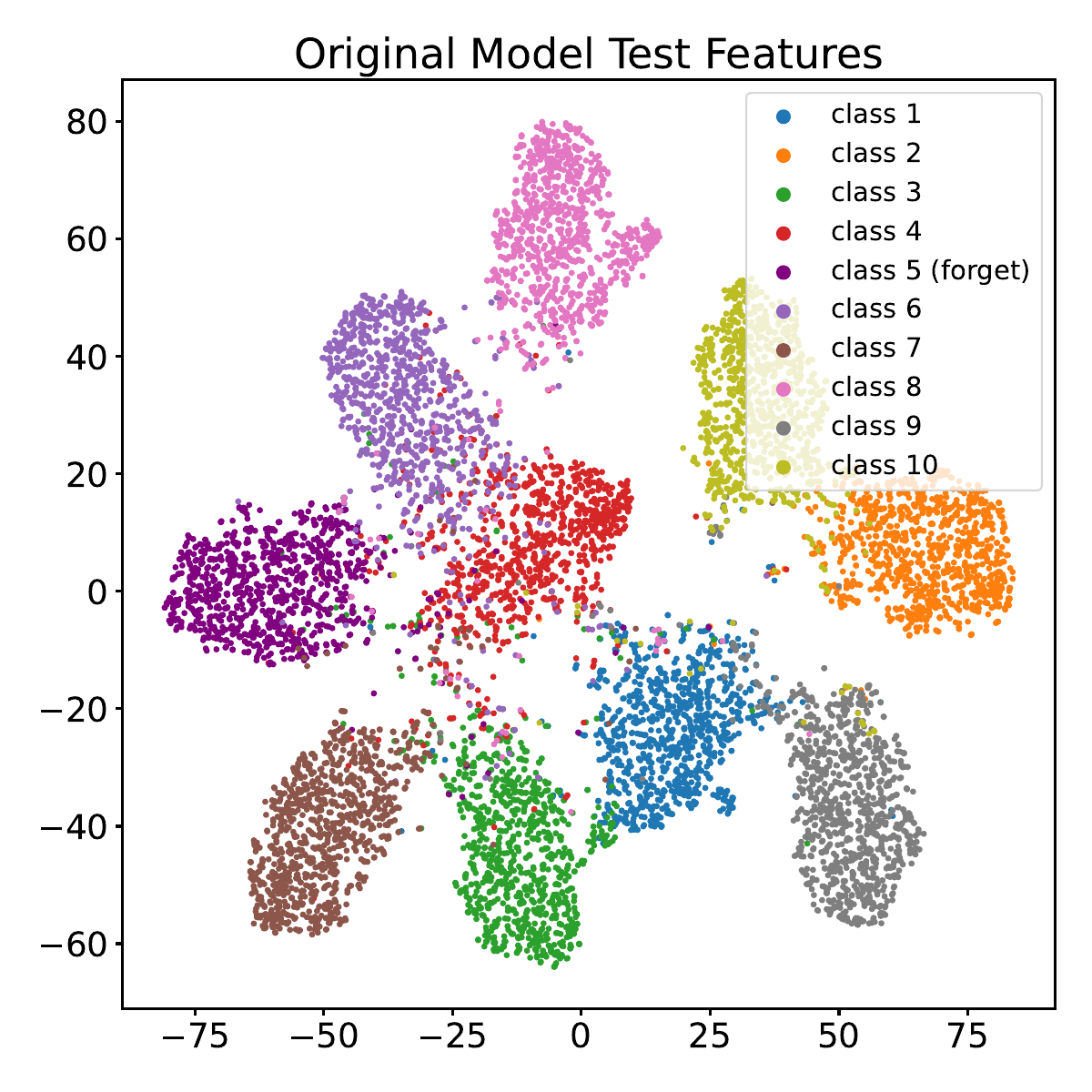}
        \subcaption{Original} \label{fig:tsne_original}
    \end{minipage}%
    \hspace{0.02\linewidth}
    \begin{minipage}{0.4\linewidth}
        \centering
        \includegraphics[width=\linewidth]{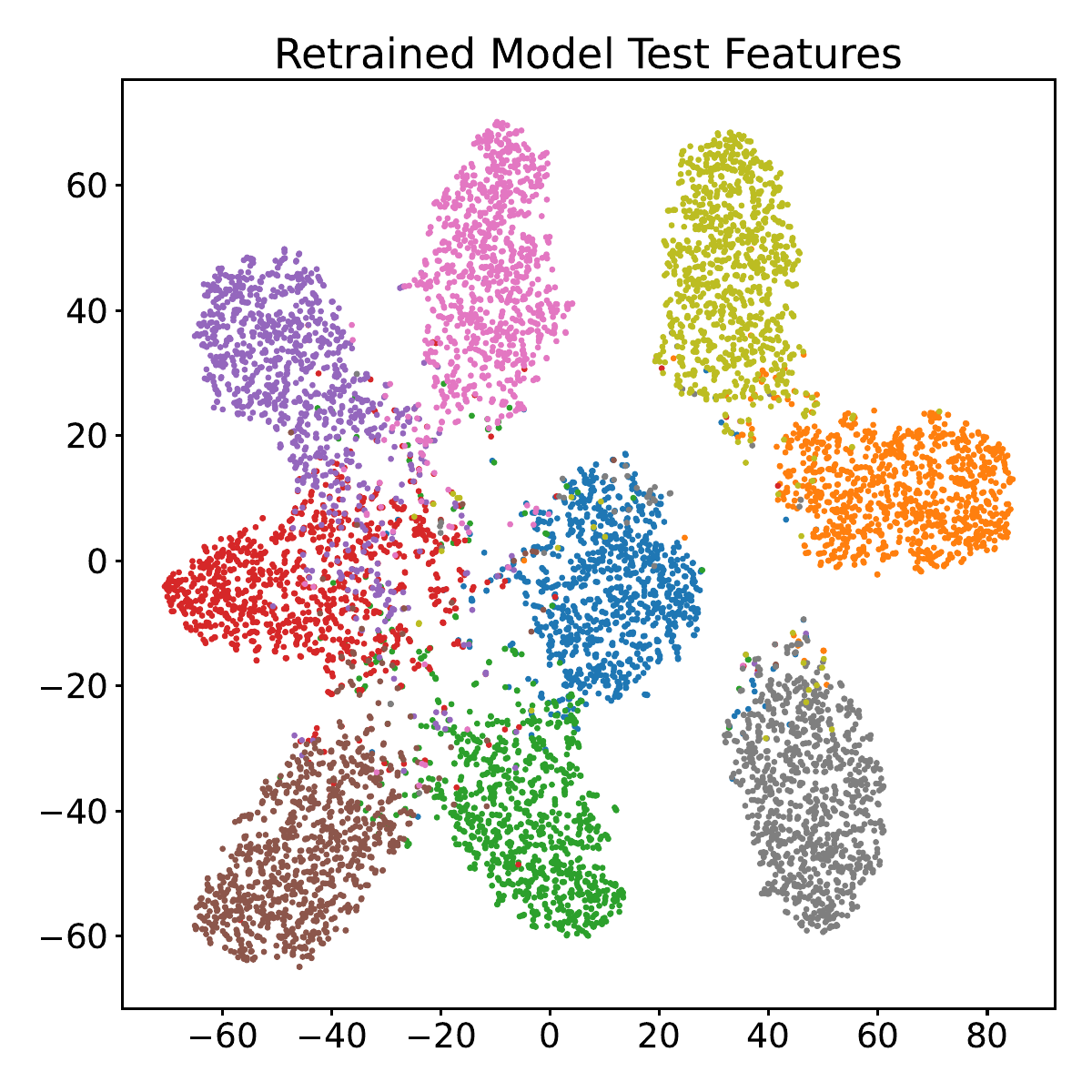}
        \subcaption{Retrain} \label{fig:tsne_retrain}
    \end{minipage}
    
    \vspace{0.2cm} 

    \begin{minipage}{0.3\linewidth}
        \centering
        \includegraphics[width=\linewidth]{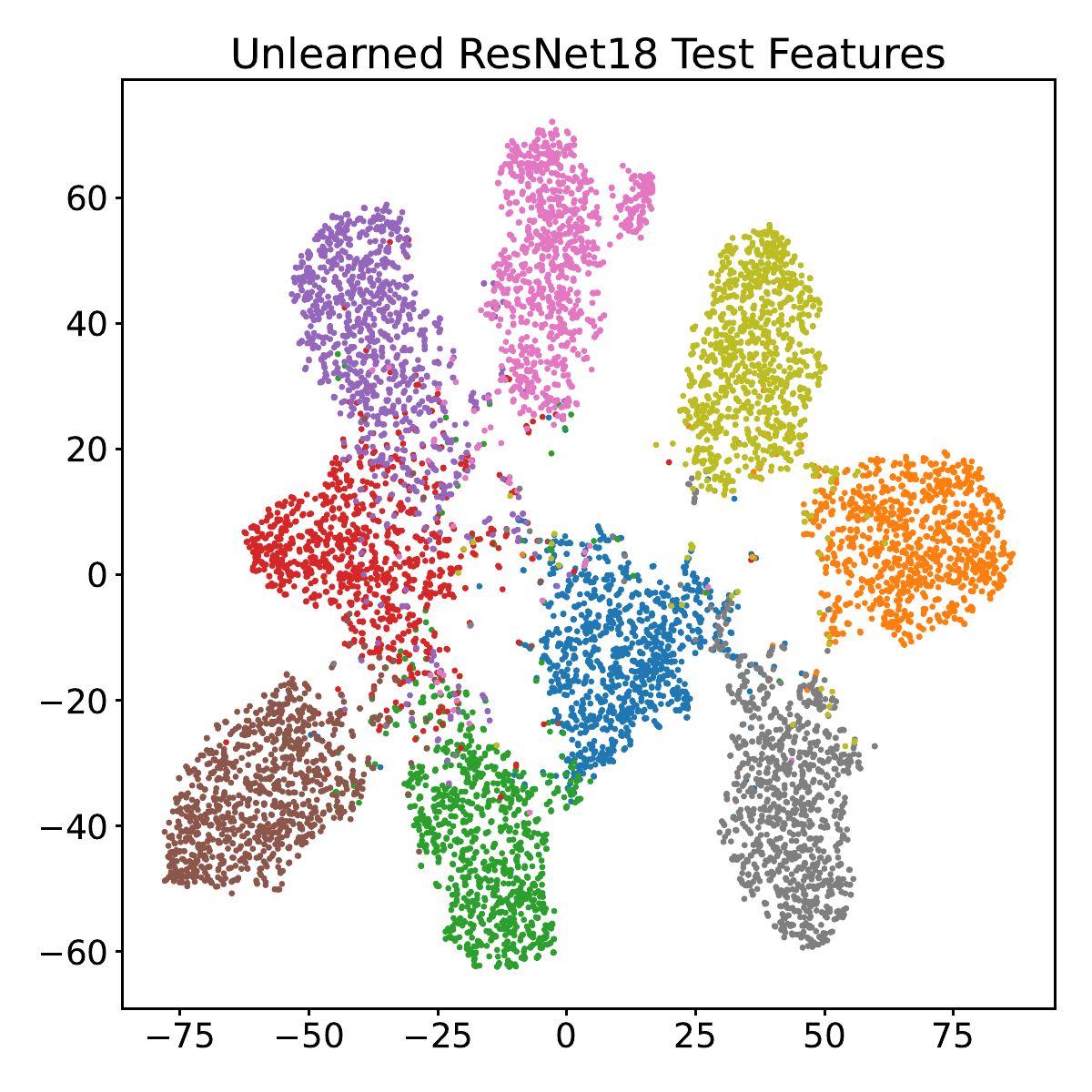}
        \subcaption{PTE} \label{tsne_pte_unlearn}
    \end{minipage}%
    \hspace{0.02\linewidth}
    \begin{minipage}{0.3\linewidth}
        \centering
        \includegraphics[width=\linewidth]{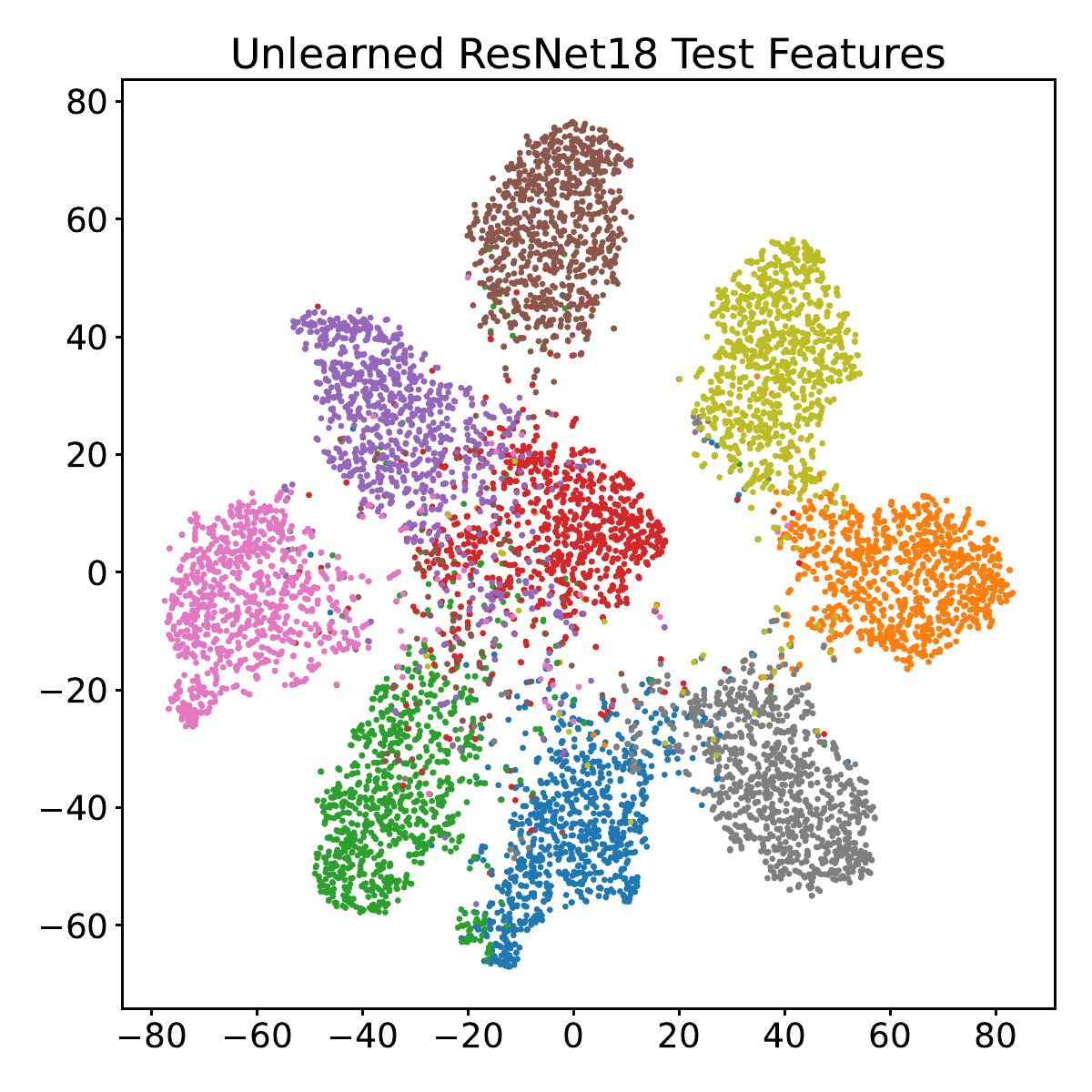}
        \subcaption{Boundary Shrink} \label{fig:tsne_boundary_shrink_unlearn}
    \end{minipage}%
    \hspace{0.02\linewidth}
    \begin{minipage}{0.3\linewidth}
        \centering
        \includegraphics[width=\linewidth]{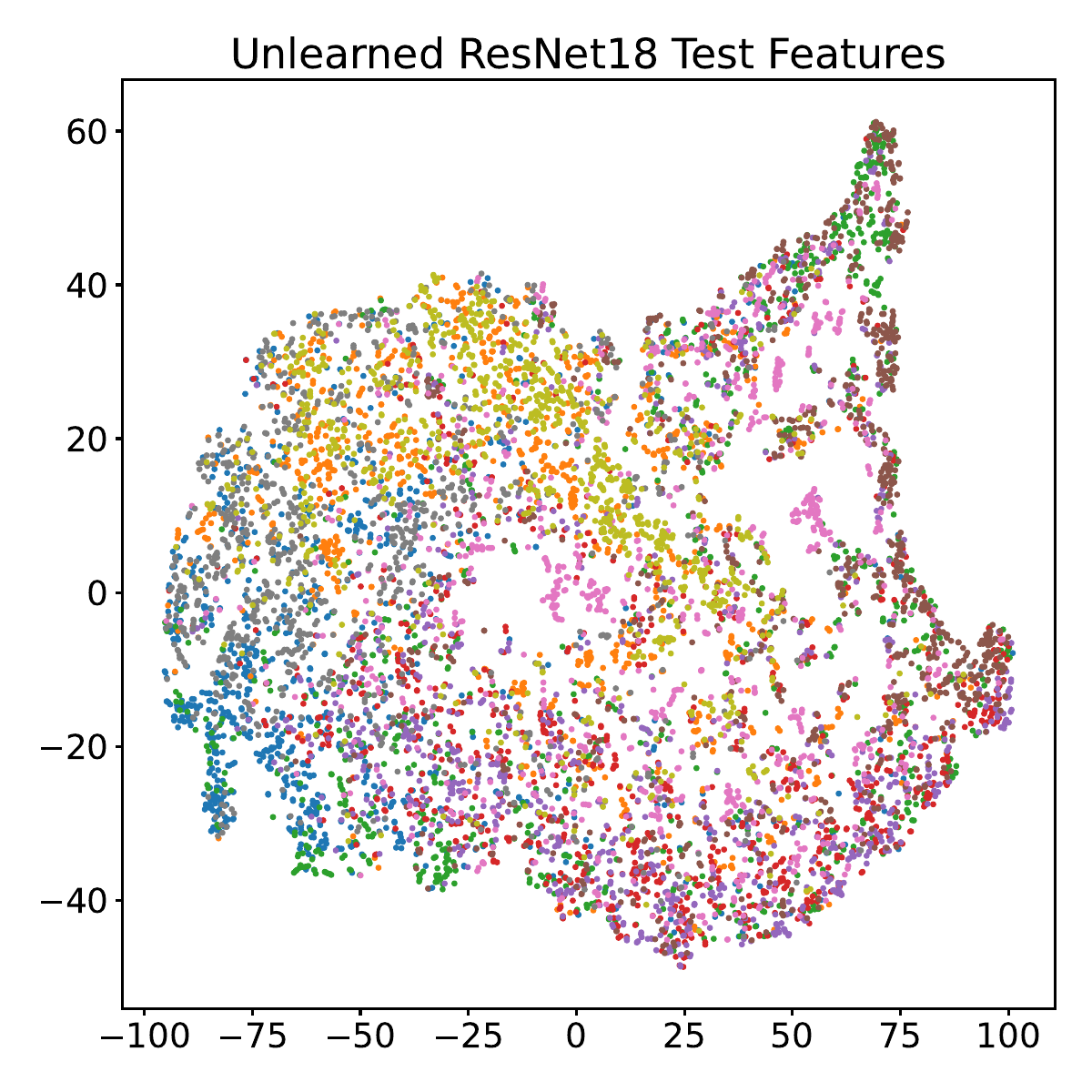}
        \subcaption{Gradient Ascent} \label{fig:tsne_gradient_ascent_unlearn}
    \end{minipage}
    \caption{T-SNE visualization of feature representations space.}
    \label{fig:1x5}
\end{figure}

To further analyze the impact of forgetting on the model’s feature space, we use the CWRU dataset and apply t-SNE to visualize the feature distribution, with different colors representing different predicted classes. Figure~\ref{fig:tsne_original} shows the feature distribution of the original model, indicating that the model clearly delineates the class boundaries, with deep purple marking the samples belonging to the forgotten class. Our method (Figure~\ref{tsne_pte_unlearn}) and the retrained model (Figure~\ref{fig:tsne_retrain}) both successfully forget the deep purple-marked forgotten class samples while retaining the boundaries of the other classes. In contrast, the Boundary Shrink method (Figure~\ref{fig:tsne_boundary_shrink_unlearn}) causes noticeable boundary degradation in some categories. For example, light purple and red points are interwoven and hard to distinguish. The Gradient Ascent method (Figure~\ref{fig:tsne_gradient_ascent_unlearn}) leads to severe boundary degradation, with samples from all categories scattered across different regions of the feature space, making it nearly impossible to distinguish between categories.

\section{Conclusion}

We proposed Probing then Editing (PTE), a novel retain-free machine unlearning framework designed for dynamic Industrial IoT environments. By conceptualizing unlearning as a structured ``probe-edit" process, PTE achieves selective forgetting through a push-pull optimization mechanism: the push branch actively dismantles the decision region of the forget class using self-generated editing instructions, while the pull branch preserves model utility via masked knowledge distillation without accessing retained data. Extensive experiments on general and industrial benchmarks—including CWRU and SCUT-FD—demonstrate that PTE achieves complete forgetting with minimal impact on retained knowledge, outperforming existing methods in both effectiveness and stability. The framework’s efficiency, compatibility with diverse architectures, and strong empirical performance establish PTE as a practical and scalable solution for class-level unlearning in real-world IIoT applications.

\newpage
\bibliography{references}
\end{document}